\documentclass{llncs}

\usepackage{subfigure}
\usepackage{colortbl}

\usepackage{graphicx}  
\usepackage{url}       

\usepackage{amsmath}   
\usepackage{cite}
\usepackage{fancyhdr} 
\usepackage{amsfonts,amssymb}
\usepackage{amssymb}

\usepackage{stfloats}
\usepackage{cases}
\usepackage{algorithm}
\usepackage{multirow}
\usepackage{algorithmic}
\usepackage{setspace} 

\begin{document}


\mainmatter              

%
\title{An optimized system to solve text-based CAPTCHA}
\author{Ye Wang, Mi Lu}
\tocauthor{Ye Wang, Mi Lu}
\institute{Department of Electrical and Computer Engineering\\Texas A\&M University, College Station, Texas, 77843, USA\\
\email{wangye0523@tamu.edu, mlu@ece.tamu.edu}}

\maketitle              




\begin{abstract}
CAPTCHA(Completely Automated Public Turing test to Tell Computers and Humans Apart) can be used to protect data from auto bots.  Countless kinds of CAPTCHAs are thus designed, while we most frequently utilize text-based scheme because of most convenience and user-friendly way \cite{bursztein2011text}.
Currently, various types of CAPTCHAs need corresponding segmentation to identify single character due to the numerous different segmentation ways. Our goal is to defeat the CAPTCHA, thus firstly the CAPTCHAs need to be split into character by character. There isn't a regular segmentation algorithm to obtain the divided characters in all kinds of examples, which means that we have to treat the segmentation individually.
In this paper, we build a whole system to defeat the CAPTCHAs as well as achieve state-of-the-art performance. In detail, we present our self-adaptive algorithm to segment different kinds of characters optimally, and then utilize both the existing methods and our own constructed convolutional neural network as an extra classifier. Results are provided showing how our system work well towards defeating these CAPTCHAs.
\end{abstract}


%

\section{Introduction}
CAPTCHA is used to tell human beings and computer programs apart automatically. CAPTCHA designers change the combination of coloring numbers and characters and so on which can be recognized by people but not the automated bots\cite{chellapilla2005building}\cite{ling2012case}, besides both companies and individuals would like to apply text-based CAPTCHAs most frequently because of the convenience. 

To defeat text-based CAPTCHA, three steps are normally needed: preprocessing by denoising, segmentation to get individual characters and recognition to identify each character. Those three steps are treated as equally important. Regarding preprocessing step, since there may be a lot of noise affecting the performance, we must decrease the effect to obtain the clear images for higher performance. Nowadays, a lot of methods are proposed to achieve the goal, like some image processing ways and machine learning algorithms such as median filter\cite{perreault2007median}, neighborhood filter, wavelet threshold, universal denies, K-nearest neighbors algorithm, support vector machine and so on. However, only when the appropriate denoising method is selected, we can acquire the clearest output\cite{bursztein2011text}\cite{wang2016self}.

After preprocessing, we should segment the image into characters individually. This is because if the image can be divided perfectly, it will be helpful for next step about the recognition accuracy. As we mentioned before, the CAPTCHAs' images are not always the same, thus defeating the CAPTCHAs itself heavily depends on the detailed weakness. After numerous methods about segmentation experiments, image intensity histogram and color clustering\cite{huang2010efficient} are two most effective ways. Last but not least, we present our novel adaptive algorithm to optimize the segmentation in defeating the CAPTCHAs which will be further discussed.
\begin{figure}
\centering
\includegraphics[width = 0.7\textwidth]{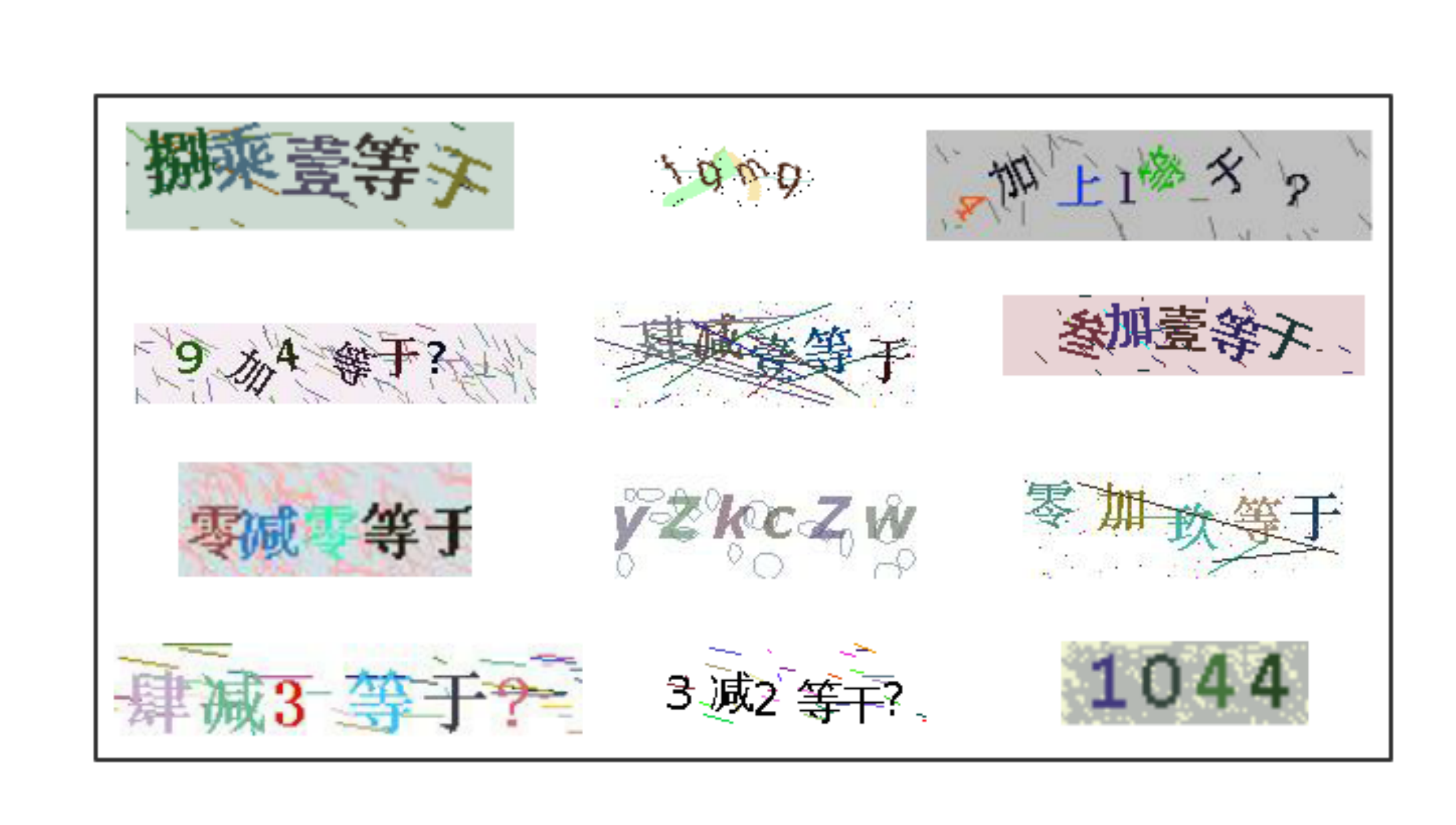}
\caption{Online samples in the datasets}\label{fig1}
\end{figure}

Recognition is the last step to get the output of defeating the CAPTCHAs. Through a lot of explorations, we conclude three fast and reliable recognition ways, they are Optical Character Recognition(OCR), Template Matching(TM) and convolutional neural network(CNN). 
First of all, OCR is used to convert words in pictures to the formal printed text\cite{smith2007overview}. Actually, OCR has been applied in a wide range of areas such as documents like contracts, passport, receipts, even business and credit cards. Besides, many commercial applications also have been developed for the task of identifying machine typed text. While OCR has failed to defeating the CAPTCHAs due to various reasons. Some commercial OCR methods can only work well on black-and-white images, part of them are even based on free-noise text. However, majority of the CAPTCHAs we need to deal with in this paper are against free-noise text, they are always random noised, colored and even rotated\cite{wang2017combining}\cite{bursztein2011text}. Moreover, the examples of our datasets suffer from various fonts, directions and even other distortions. So that the combination of those existing problems can not be easily solved by just employing OCR. 

To overcome OCR's shortage, we would like to introduce TM. Long long ago, before OCR, this is the most original machine learning way to perform recognition\cite{lewis1995fast}. TM is straightforward and easy to implement. Specifically, when we defeat some CAPTCHAs with rotated examples, TM can achieve higher accuracy rate than OCR. With the increasingly complicated circumstances designed, in terms of heavily rotation, overlapping and twisted, CNN overwhelmingly works better than any others in terms of accuracy and time complexity\cite{krizhevsky2012imagenet}. However, constructing a CNN is also huge, which need to pre-train numerous well prepared samples \cite{lawrence1997face}. Our own CNN conquers this problem with a lot of manually efforts aiming at data collection. We finally obtain the state-of-the-art performance to defeat the CAPTCHAs.

The rest of this paper is organized as follows: Section II describes the background of three steps in terms of denoising, segmentation and recognition. Section III proposes the scheme of adaptive segmentation and how we construct our convolutional neural network. Section IV shows the experiments. The paper concludes in Section V.

\section{Background}

\begin{figure}
\centering
\includegraphics[width = 0.6\textwidth]{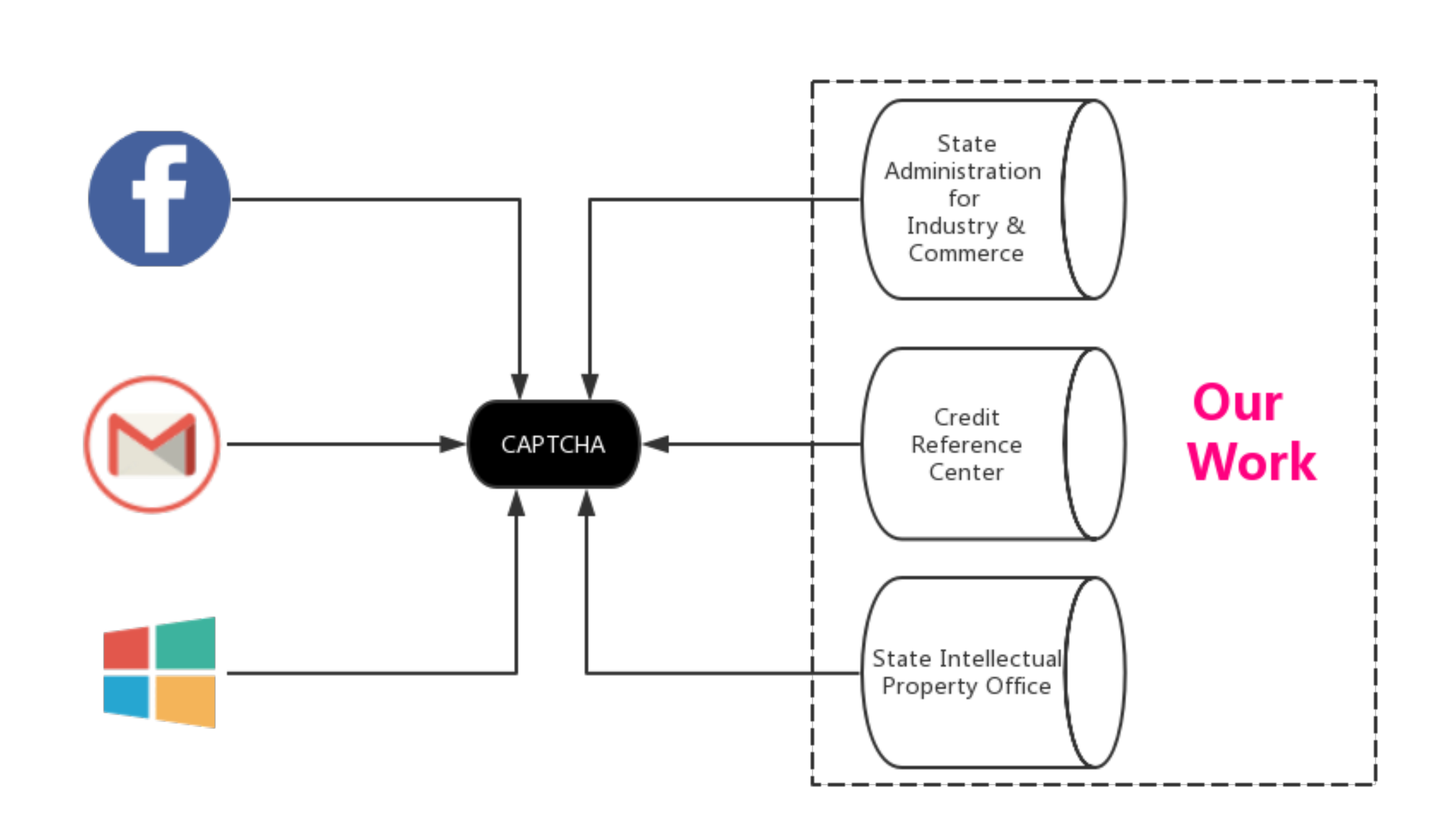}
\caption{CAPTCHA datasets}\label{fig1}
\end{figure}

CAPTCHAs can be found almost in every websites\cite{von2003captcha}\cite{ling2012case}\cite{roshanbin2013survey}. Each type of CAPTCHA should be solved accordingly because of the unique encoding algorithms. Correspondingly, a lot of deCAPTCHA algorithms have been proposed for most famous companies such as Google, Microsoft, and Facebook. A low-cost attack on a Microsoft CAPTCHA was presented for solving the segmentation resistant task, which achieved a segmentation success rate of higher than 90\%\cite{yan2008low}. Moreover, another projection-based segmentation algorithm for breaking MSN and YAHOO CAPTCHAs proves to be effective, which doubled the corrected segmentation rate over the traditional method\cite{huang2008projection}. Besides, an algorithm using ellipse-shaped blobs detection for breaking Facebook CAPTCHA also presents to be useful\cite{liu2013efficient}.

As can be seen, Fig. 1 presents some examples of text-based CAPTCHAs which will be discussed later. Fig. 2 describes the system of our datasets. Fig. 3 shows the basic flow chart to defeat the CAPTCHAs. In this section, we will demonstrate the whole process.

\begin{figure}
\centering
\includegraphics[width = 0.6\textwidth]{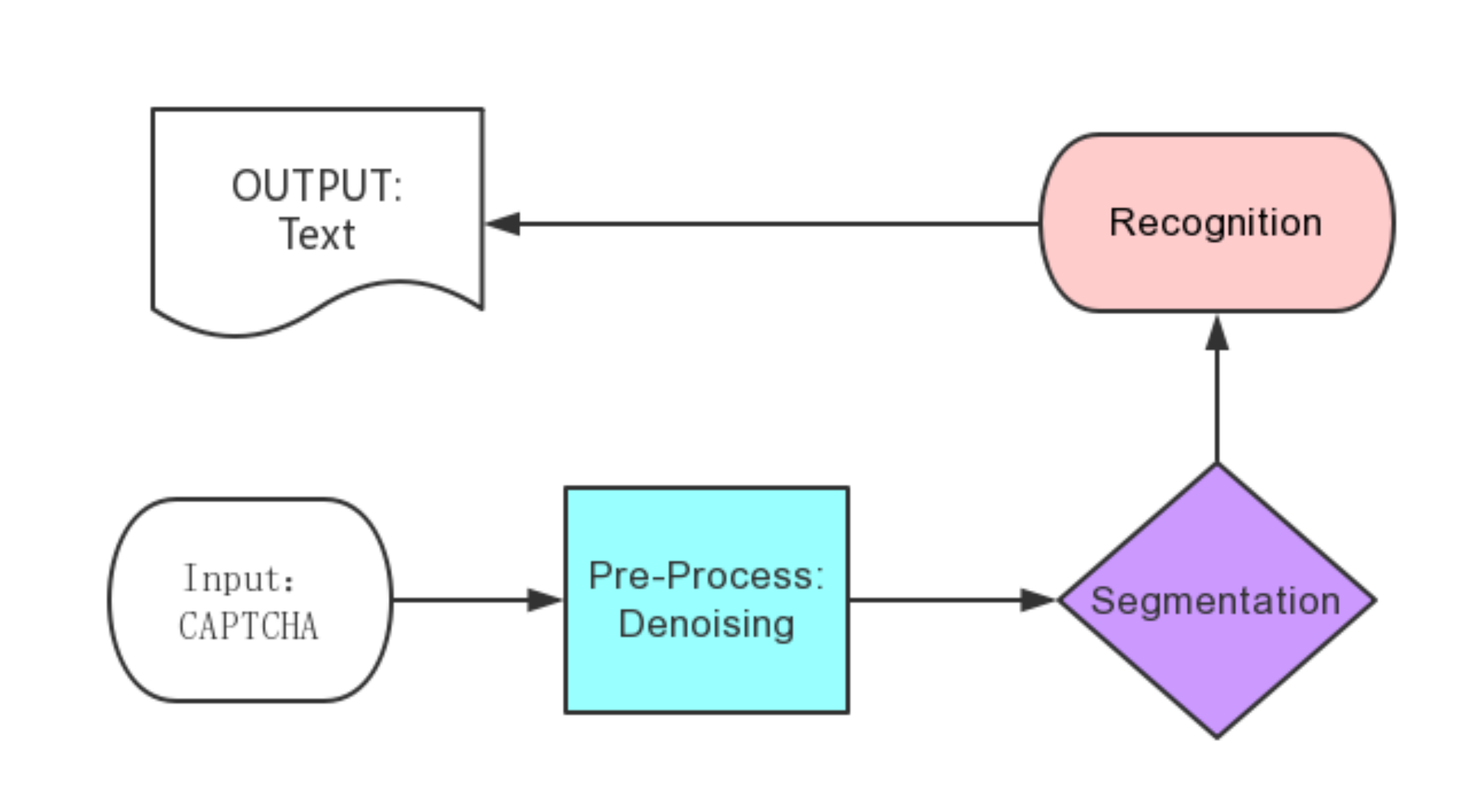}
\caption{Basic Flow Chart to defeat the CAPTCHA}\label{fig2}
\end{figure}

\subsection{Prepossessing Step}
There are several possible techniques to denoise the images based on various types of each other. After our efforts, we find out the appropriate ways according to the different types of examples. Median filter is effective for nonlinear digital method to filter the noise. Affine transformation acts an important role in preprocessing as well. Actually affine transformation does not necessarily preserve angles between lines or distances between points, though it does preserve ratios of distances between points lying on a straight line\cite{stearns1995method}. After an affine transformation, the sets of parallel lines can still remain parallel, which also preserve points, lines and planes.
	
\subsubsection{Thresholding} In an image, the main difficulty of finding the optimal intensity of the whole image between real image and noise is thresholding, because there are not any relationships between the pixels. No one can guarantee that the pixels between each others identified by the thresholding process are contiguous. We can easily include irrelative pixels that aren't part of the desired region in real image we actually need, and we can easily miss isolated pixels within the region as well (especially near the boundaries of the region). These effects get increasingly worse as the type of noise becomes more and more complicated, simply because it's more likely that a pixel intensity cannot represent free-noise image intensity in the region \cite{rashidi2012implementation}. When we use thresholding method, we typically have to balance with the tradeoff in terms of losing too much of the region informations and getting too many background pixels with noise. (Shadows of objects in the image are also a real pain - not just where they fall across another object but where they mistakenly get included as part of a dark object on a light background.) 

Right here, we utilize the Automated Methods for Finding Thresholds: To set a global threshold or to adapt a local threshold to an area, we usually look at the histogram to see if we can find two or more distinct modes, one for the foreground and one for the background\cite{kurugollu2001color}.

\subsubsection{Median filter}Median filter is effective for nonlinear digital signal to remove noise as well as preserve edges while removing noise. Noise reduction is a typical pre-processing step to improve the final results of later processing (for example, edge detection on an image). 
The main idea of the median filter is to run through the signal entry by entry, replacing each entry with the median of neighboring entries. For example, for every window slides, y[1] = Median[2 2 80] = 2. Here below is shown one classic algorithm \cite{perreault2007median} .

\begin{algorithm}
  \label{alg:1}
  \caption{Median Filtering Algorithm}
 \begin{algorithmic}
  \REQUIRE  {Image $\boldsymbol{X}$ of size m $\times$ n, kernel radius $\tau$}
  \ENSURE Image $\boldsymbol{Y}$ of the same size as $\boldsymbol{X}$

  \STATE \textbf{Initialize:}  \text{kernal histogram $\boldsymbol{H}$}

\STATE \textbf{for}  i = 1 to m  \textbf{do}
   \STATE \qquad  \textbf{for}  j = 1 to n  \textbf{do}
    \STATE \qquad \qquad  \textbf{for}  k = -$\tau$ to $\tau$  \textbf{do}
\STATE \qquad \qquad \qquad Remove $\boldsymbol{X}_{i+k,j- \tau-1}$ from $\boldsymbol{H}$
\STATE \qquad \qquad \qquad Add $\boldsymbol{X}_{i+k,j+ \tau}$ to $\boldsymbol{H}$
    \STATE \qquad \qquad  \textbf{end for}  
    \STATE \qquad \qquad   {$\boldsymbol{Y}_{i, j}$} $\gets$   $median(\boldsymbol{H})$
\STATE \qquad \textbf{end for} 
\STATE  \textbf{end for} 
\end{algorithmic}
\end{algorithm}

\subsubsection{Affine transformation}In geometry, an affine transformation, affine map or an affinity is a function between affine spaces which preserves points, straight lines and planes. An affine transformation may not necessarily preserve angles between lines or distances between points, though it does preserve ratios of distances between points lying on a straight line\cite{stearns1995method}.

\begin{figure}
\centering
\includegraphics[width = 0.45\textwidth]{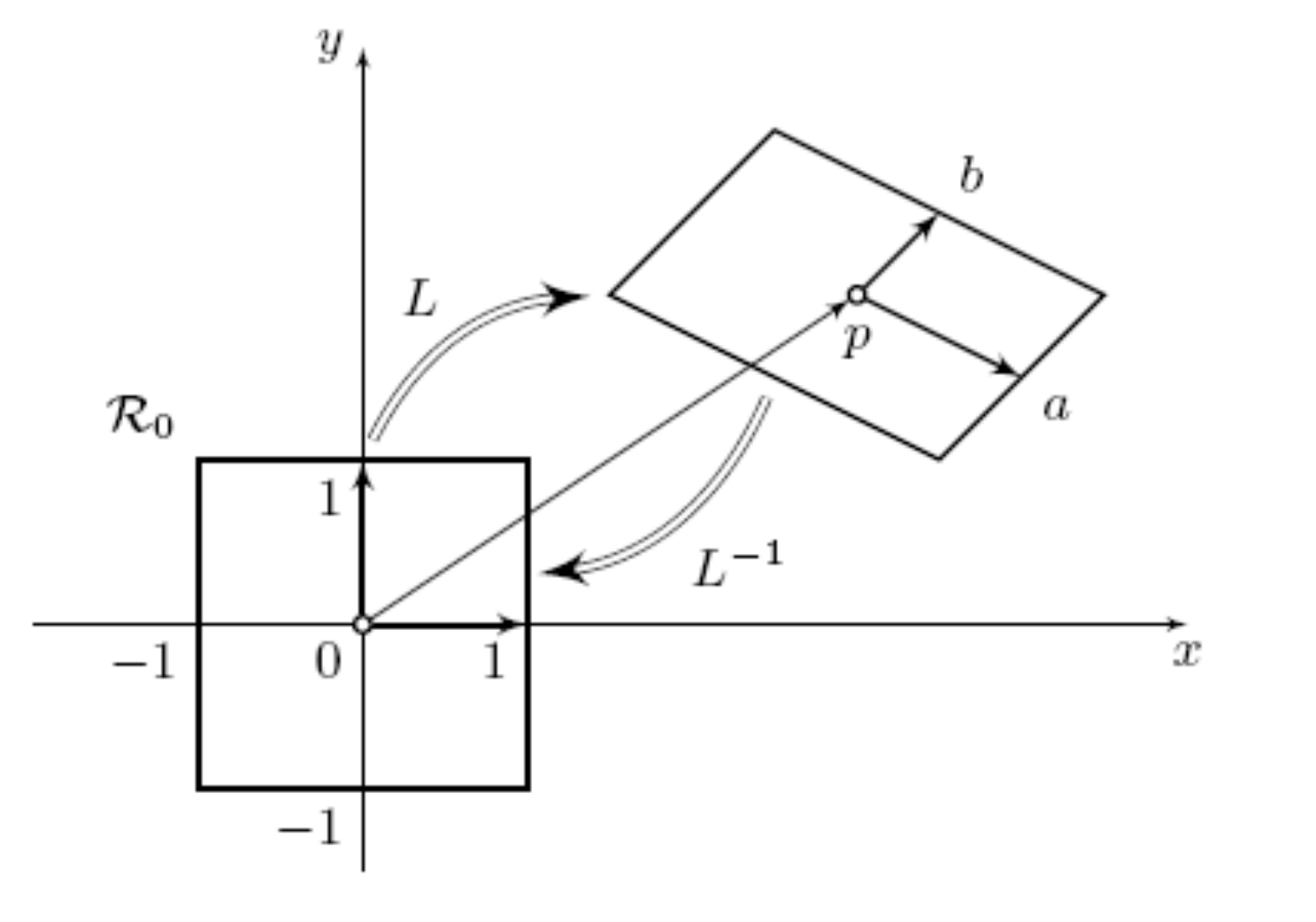}
\caption{A classic example of affine transformation}\label{fig4}
\end{figure}

An affine map is made up of two functions: a translation and a linear map. The ordinary vector algebra represents linear maps by matrix multiplication, and represents translations by vector addition. If the linear map is expressed as a multiplication by a matrix A and the translation as the addition of a vector $\vec{b}$, an affine map f acting on a vector $\vec{x}$ can be represented as
\begin{equation}
\vec{y} = f(\vec{x}) = A\vec{x} + \vec{b}
\end{equation}

In Fig 4, we can easily understand the effect of affine transformation. In short, we need to retain the ratio among graphics while the angles cannot be guaranteed.

\subsection{Segmentation} Segmentation is quite important, which can heavily influence the performance. The most common way for analyzing segmentation is to divide the CAPTCHAs' image into multiple single parts\cite{huang2010efficient}\cite{zhang2016effective}. Segmentation is used to detect the examples of CAPTCHAs' boundaries like lines or curves, with consideration of the rotation, noise and even twisted characters.

Histogram-based and K-means clustering are two effective methods in segmentation. Histogram-based is to count the quantity of pixels in each row or column in grey level. The basic algorithm would be illustrated as $y_{0}$, $y_{1}$, ..... $y_{n}$, where $y_{i}$ is the number of pixels in the image with gray-level $i$, and $n$ is the maximum gray-level attained. Imagine that if the distance in histogram between each character is very far, it would be easy to separate \cite{glasbey1993analysis}. While K-means clustering is another method for segmentation. Actually, segmentation is a very dependent method since the CAPTCHAs vary considerably. So far, there aren't any useful algorithms to segment the affixed, bended, even twisted characters. There are several generations of K-means methods. Following is the basic pseudocode we would need to exploit.
In the next section, we will introduce our adaptive length system with different kinds of CAPTCHAs.

\begin{algorithm}
  \label{alg:2}
  \caption{K-Means Algorithm}
 \begin{algorithmic}

\STATE  \text{$\boldsymbol{1.Select}$ K points as the initial centroids} 
\STATE  \text{$\boldsymbol{2.Repeat}$} 
 \STATE\qquad  {$\boldsymbol{Form}$ K clusters by assigning all points to the closest centroid}
 \STATE\qquad  {$\boldsymbol{Recompute}$ the centroid of each cluster} 
\STATE  \text{$\boldsymbol{3.Until}$ The centroids don't change} 
   
  \end{algorithmic}
\end{algorithm}

In the next section, we will focus on an algorithm which contains a lot of CAPTCHAs by using adaptive length segmentation.

\subsection{Recognition} Last step is to recognize the characters automatically, getting printed message to fulfill the whole process of defeating the CAPTCHAs.
We implement the recognition in three ways, Optical character recognition(OCR), Template Matching(TM) and Convolutional Neural Network(CNN).
For OCR, we choose an open source software named Tesseract by Google \cite{smith2007overview} \cite{smith2009adapting}. If the format is rigid, the recognition accuracy performance would be the best. While not all the CAPTCHAs can be similar, for some irregular cases, TM is an option, which is a pattern-oriented method to find the most similar candidate characters. The theory is to slide the template image over the input image, then get the similarity matrix of each other to obtain the best candidate.

CNN is a more complicated artificial intelligent technique compared with OCR and TM\cite{goodfellow2013multi}\cite{krizhevsky2012imagenet}. Since it needs much more pre-defined data to train the classifier for a system in terms of higher accuracy rate and less time complexity. This is because the more data we can use, the better classifier we would get. The limitation of CNN is also apparent in terms of the size of datasets, since we need to manually denoise the image first and then segment them into individual character, thus the working procedure is comparably complicated than the others. 

\section{The scheme}
\subsection{Denoise Filter}There are many types of CAPTCHA examples and each type of them is not the same as others. Generally, we need to first denoise the examples for clear images as we mentioned before. There are two major denoise methods regarding image processing, one is to carry on in frequency domain and the other is to process in time domain. In the frequency domain, we usually transfer the image from time series to frequency series, utilizing a corresponding transformation such as Fourier Transformation(FT) which can reflect image characteristics to denoise the image. In FT spectrum, the noise always exists around high frequency areas while the image entity itself exists in low frequency areas. So some particular low pass filters can be adopted to eliminate the noise and to retain the filtered real image. Normally, the filters are divided into linear and non-linear one for denoising. Actually, linear filters such as mean filter and gaussian filter can be considered as a result of raw data and the filters in arithmetic computations like addition, subtraction, multipllication and division. Due to the limitation of arithmetic computation in linear filters, the transformation function is determinate and unique. 

In the time domain, compared with the linear filter, the process of non-linear filtering can be viewed as passing raw data though specific filters such as minimum filter, maximum filter and median filter. As to the implementation, nonlinear filters act like to slide a window to make the decision about minimum, maximum or median value within the window frame. Unlike linear filters, this operation is nondeterminate because of the logic relationship in window sliding. What's more, compared with linear filters, the function of nonlinear filters is not limited to protecting the margins but also removing the noise much better. However, every coin has two sides. The speed in non-linear filters would be slower than that in linear filters, because we have to slide the window over the whole image. Median filter is a typical nonlinear technique. The basic idea is to get the median value of a pixel in grey level within the window. Since the noise pixels are always higher or lower than the image entity itself, this method can perform significantly in denoising and margin protection.

\subsection{Adaptive algorithm} There are various kinds of similar notations defeating CAPTCHAs. As an example, we invested the popular query involving the following triplet structure,$<$ first operand, operator, second operand$>$. 

One may need to know that the length of operator is variant here. Also, the operands can be of different fonts, like traditional Chinese characters, simplified Chinese characters and Arabic numbers. Furthermore, the font of the operator is orthogonal to that of the operands.
\begin{figure}
\centering
\includegraphics[width = 0.7\textwidth]{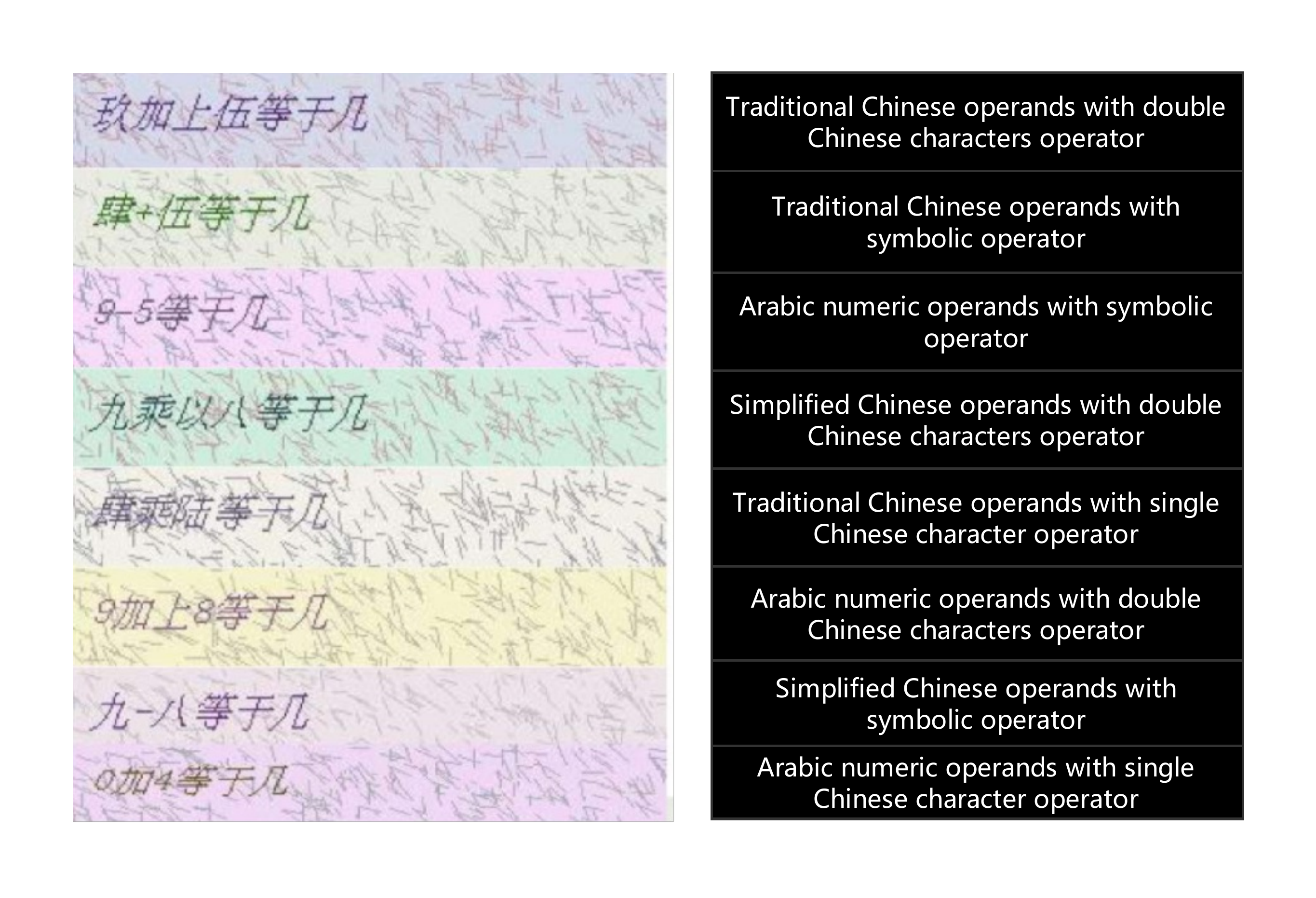}
\caption{The example of adaptive segmentation}\label{fig5}
\end{figure}
There are three kinds of operators, symbol operator, single character and double characters operators, each of them also includes several types of operations. The ones given in Fig. 5 show the challenge involving Chinese characters. For example, in the first row, it asked you to answer the question:$9+5=?$. The first operand and second operand are traditional Chinese characters, and the operator is made up of two characters. While in the third row, there is a symbol operator with Arabic number operands which means $9-5=?$.

Let's start with the operand first. The first operand can be made up with any digits, so can the second operand. We assume that the number of the first operands, the second operands and the operator are M, N and O respectively. Since they are pairwise orthogonal, the total number of combinations of those notations can be up to $M*N*O$ which is very massive. Fortunately, in our cases, some rules can be applied to reduce the number of possible combinations. 

For example, given one operand with values zero to nine to represent: there are three fonts like traditional Chinese characters, simplified Chinese characters and Arabic numbers, resulting in $10*3=30$ potential choices in total.

\begin{figure*}
\centering
\includegraphics[width = 0.85\textwidth]{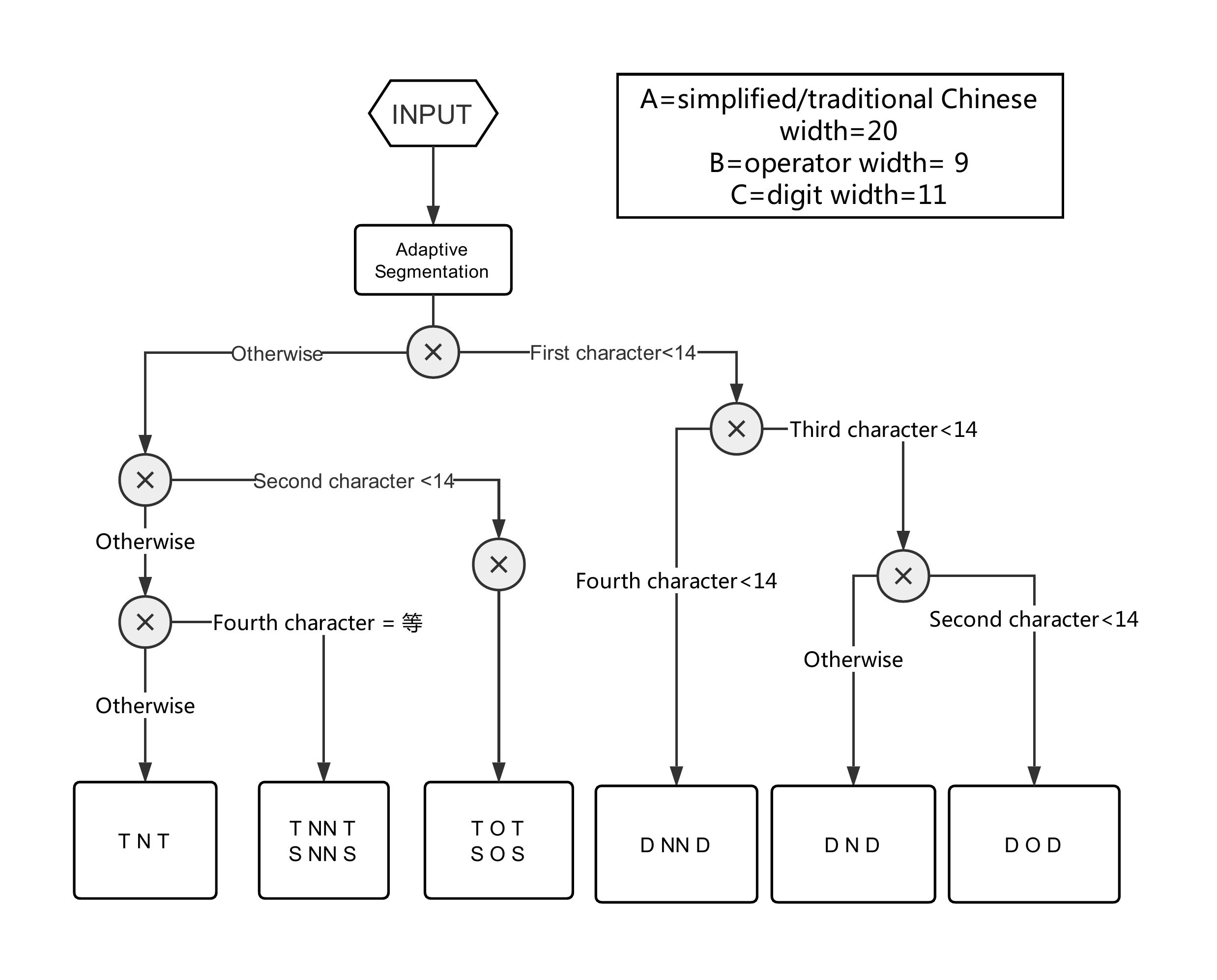}
\caption{The detailed flow chart of adaptive segmentation}\label{fig6}
\end{figure*}

Moreover, the operators representing addition, subtraction, multiplication or division can be of three fonts, symbolic operators, single Chinese characters and double Chinese characters. The total number of operator candidates is $4+4+4=12$. With the two operands and one operator taken into consideration, the total number of combinations is $30*30*12=10800$. 

\begin{table}[hb]
\centering
\begin{tabular}{|c|c|c|c|c|c|c|c|c|c|c|}
\hline
\multicolumn{3}{|c|}{Type 1} &  & \multicolumn{3}{c|}{Type 2} &  & \multicolumn{3}{c|}{Type 3} \\ \hline
\textbf{S} & \textbf{O} & \textbf{S} & \textbf{} & \textbf{T} & \textbf{O} & \textbf{T} & \textbf{} & \textbf{D} & \textbf{O} & \textbf{D} \\ \cline{1-3} \cline{5-7} \cline{9-11} 
\textbf{S} & \textbf{N} & \textbf{S} & \textbf{} & \textbf{T} & \textbf{N} & \textbf{T} & \textbf{} & \textbf{D} & \textbf{N} & \textbf{D} \\ \cline{1-3} \cline{5-7} \cline{9-11} 
\textbf{S} & \textbf{NN} & \textbf{S} & \textbf{} & \textbf{T} & \textbf{NN} & \textbf{T} & \textbf{} & \textbf{D} & \textbf{NN} & \textbf{D} \\ \hline
\end{tabular}
\caption{All possible cases}
\label{table_1}
\end{table}

As the last but not the least important issue, Table 1 shows all the possible combinations in our cases. $S$ represents simplified Chinese, $T$ is the traditional Chinese, $D$ stands for Arabic numbers. Regarding the operator, we use $O$ for symbolic operator, and $N$ and $NN$ for the single and double Chinese characters symbols respectively. For all the possible combinations, please reference to Table 1. We divide all the combinations into three types based on the different fonts of operands. The simplified Chinese operands belong to Type 1, traditional Chinese operands belong to Type 2 and Arabic numeric operands belong to Type 3 respectively. 

Since the operators needs be paired with different kinds of operands, the combinations of operand and operator examples are enumerated in Table 1. Nine groups among the operands and operators in total are shown in Table 1. For example, single character operator can be paired with traditional Chinese operands, simplified Chinese operands and Arabic numbers. One thing needs to be mentioned here. According to the shortage of our datasets, one type of notation is missing, $S N S$ in the second row of Type 1. So we have, actually, only eight kinds of notations as shown in Fig. 5.

As we introduced before, the segmentation for each type of CAPTCHAs is a very critical operation. Since there are eight types of notations, we have to try eight times to identify the exact font type if using the traditional sequential method. We propose our algorithm to identify the font type, with the higher speedup and accuracy.

After preprocessing, we cannot directly employ the histogram-based method to apply segmentation because of the characters' rotation. Besides, the neighboring characters will affect close-by characters of each others. Therefore we have to use Affine Transformation to turn the rotated characters into vertical. Following is the representation:
\begin{eqnarray}
\left[
  \begin{array}{c}
    {x}' \\
    {y}' \\
  \end{array}
\right]=\left[
  \begin{array}{cccc}
   ax+by\\dx+ey
   \end{array}
\right]=
\left[
  \begin{array}{cc}
    a & b\\d&e
  \end{array}
\right]
\left[
  \begin{array}{c}
    {x}\\
    {y} \\
  \end{array}
\right]
\end{eqnarray}

One attractive feature of this matrix representation is that we can use it to factor a complex transformation into a set of simpler transformations. Because the rotation transformation is utilized, the following rotation matrix is essential: 
$\left[
  \begin{array}{ccc}
   {\cos\theta}&{- \sin \theta }\\ {\sin \theta} & {\cos \theta}
  \end{array}
  \right]$, where $\theta$ is an angle of counterclockwise rotation around the origin.

Whenever loading an image we first perform the regular affine transformation to make it vertical. The good thing in our application is that the rotated angels of the characters in CAPTCHA examples are fixed, so our task would be much simpler. After the optimization of the overall procedure, the whole process is exhibited in Fig. 6. We just need to check the first four characters, because the citation we deal with is the operation of the operands. We need the first operand, the operator and the second operand where the operator includes one or two characters. Then we need to identify those operators, checking the first four characters are sufficient. 

Once the notation comes into our system, we check the first character. If it is an Arabic number, we know that the notation belongs to Type 3 in Table 1, possibly in $<$D - -$>$ form. Next we goto the third character where the second operand is possibly located. If that character is a D, we are sure that $<$D O D$>$ or $<$D N D$>$ is the form of the notation. We can then go back to check the second character to confirm whether it is an operator represented in N or O. If the third character is not an Arabic number, indicated is that it is not the second operand, because the second and first operand should be of the same type. The option left is that the operator between the two operands is of NN form. In that case, the second operand must be located in the position of the fourth character. Since the first and the second operands must be of the same type as we mentioned before, the fourth character will be an Arabic number, then the notation is represented in $<$D NN D$>$ form.

If the first character is not an Arabic number, the notation could belong to Type 1 or Type 2 listed in Table 1 in $<$S - S$>$ or $<$T - T$>$ form. Next we go to the second character to verify whether it is a symbolic operator or not. If it is, the notation is then represented in $<$S O S$>$ or $<$T O T$>$ form. If the second character is not symbolic, then it can be a Chinese character, representing the operator by itself. The other case is that the operator is represented by double Chinese characters. Let's move to the third character to check. For the former case, the third character is where the second operand stands, and the second operand should have the same style as the first one. So we get $<$S N S$>$ or $<$T N T$>$ as listed in Table 1. For the later case, operator being double Chinese characters case, NN will extend to the third character position. The simplest way is to directly check the fourth character, to see whether it is equal to one particular ideograph or not.

We take the third row in Fig. 5 as an example, the meaning of which is $9-5=?$. The first character is a digit such as D, then we move to check the third one. The third character is a digit like the first operand. Then as the last examination, the second character is to be verified to confirm the operator type. It turns out that the operator is a symbolic operator, so we conclude that the notation type is $<$D O D$>$.

We look at another example next, from the 6th row of Fig. 5. We check the first character and find that it is a digit. We move to the third one and find that it is not a digit but a Chinese character. Quickly we understand that the operator is a double-character operator and the notation is in $<$D NN D$>$ form. Actually we don't have to examine the fourth character because it must be the the same type operand as the first one. We just double checked here. 

In Fig.6, we check a character by its width instead of its content, which is much faster and more reliable. We found that the width of both the simplified Chinese characters and the traditional Chinese characters are 20. Besides, the width of a symbolic operator is 9 and that of a digit is 11. We set the threshold of the width as 14.

\begin{figure}
\centering
\includegraphics[width = 0.7\textwidth]{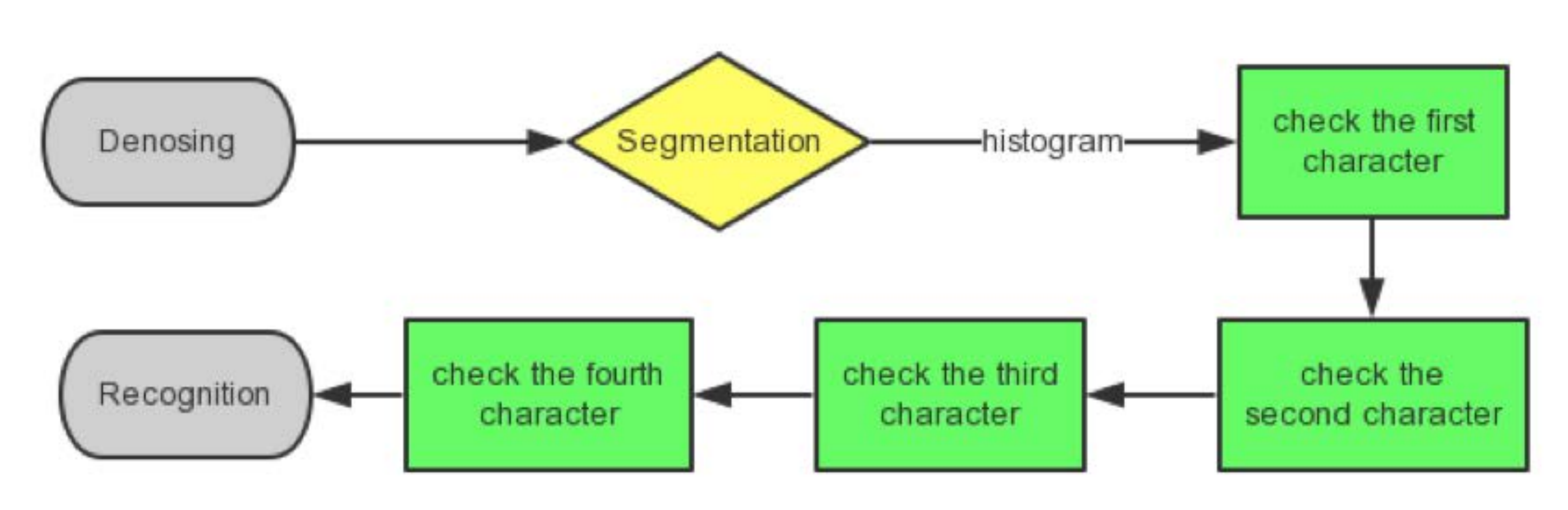}
\caption{Traditional flow chart of segmentation}\label{fig7}
\end{figure}
Fig. 7 shows the basic flow chart of segmentation which is a basic sequential test procedure. Fig. 8 illustrates the traditional serialized method. If we use the serialized segmentation as in Fig. 8, the total time to identify the exact types shown in Fig. 5 is: $T1=8*4*t$, where $t$ is the time to check a single character. 
Regarding our algorithm, as Fig. 6 illustrated, the whole procedure is optimized into at most three steps, thus the time to identify the exact type is $T2=3*t$.  
The speedup between ours and serialized segmentation is:
\[T1/T2=10.67\]

We denote the single character recognition rate as $r$. In serialized segmentation, the successive recognition rate of the whole eight types $R1$ would be:
\[R1=r^{4*8}\]
In our algorithm, the successive recognition rate $R2$ is:
\[R2=r^{3*8}\]

Ideally, compared with the traditional algorithm, the improvement we obtained is 
\[R1/R2=r\]

One can reference to Section IV, Experiment, in the later part of this paper to find the improvement of the single character recognition rate of our algorithm over the traditional algorithm, which is roughly from 75\% to 99\% depending on the exactly type of CAPTCHA.

\begin{figure}
\centering
\includegraphics[width = 0.4\textwidth]{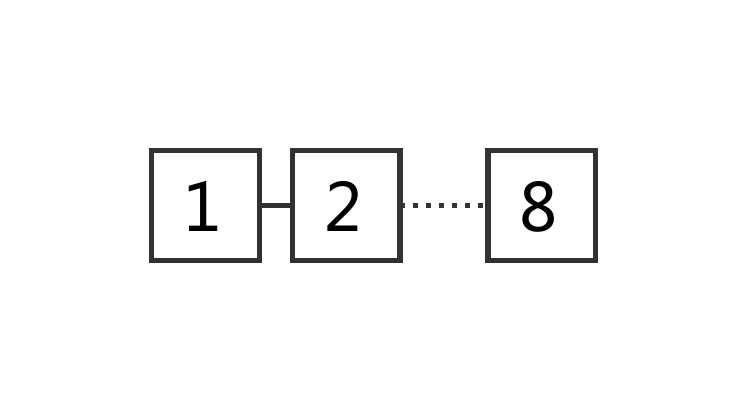}
\caption{Sequentially defeating the CAPTCHA}\label{fig8}
\end{figure}

\subsection{Convolutional neural network construction}
\subsubsection{The architecture}The CNN architecture is shown in Fig. 9, which is adapted with one of the most classic LeNet-5 \cite{lecun1998gradient}\cite{rumelhart1988learning}. In Fig. 9, the top is our image input layer and the image size is 32x32. The net contains six layers: the first four layers are convolutional layers and the remaining two are fully connected layers. The general strategy of CNN is to extract simple features at a higher resolution by the first convolutional layer, and then convert the features in higher resolution into more complex features at a coarser resolution\cite{simard2003best}. To obtain the coarser resolution, the most frequently used method is to sub-sample a layer by a factor of two, which can be utilized for the size of convolutional kernel as shown in the second layer in Fig. 9. Regarding the kernel's width, the criteria is to have enough overlap to keep useful information, as well as not to have too much redundant computation. Thus we choose five by five as the kernel size because three by three is too small with only one unit overlapping and seven by seven is too big, with 5 units or over 70\% overlapping. After the first convolutional layer extracting features, we found that if the kernel size is 5, the overall performance is better than the case with a kernel size less than 5, while increasing the kernel size to bigger cannot improve the performance too much. Then we sub-sample those features to decease the data size, prevent over-fitting as well as preserve the important information such as margins. Similarly we repeat the first and second layer to construct the third and fourth layer, so that we can carry enough information to the classification layer. Finally we add a trainable classifier to the extractor which is of two fully connected layers. The output of the last fully connected layers is fed to a 82-way (10 for digits 0-9, 26 for lower case letters a-z and 26 for capital letters A-Z, 10 for simplified Chinese digits and 10 for traditional Chinese digits) softmax which produces a distribution over 82 class labels. 

\subsubsection{Pooling layer}Pooling layers in CNNs summarize the outputs of neighboring groups of neurons in the same kernel map. Traditionally, the neighborhoods summarized by adjacent pooling units do not overlap\cite{jaderberg2014synthetic}. A pooling layer is to construct a grid of pooling units spaced $s$ pixels apart, each summarizing a neighborhood of size z x z centered at the location of the pooling unit. Regarding the size of pooling size, typical values are 2x2 or no max-pooling. Very large input images may warrant 4x4 pooling in the lower-layers. However, such a large 4x4 pooling layer will reduce the dimension of the signal by a factor of 16, and may cause throwing away too much information.\cite{krizhevsky2012imagenet}\cite{goodfellow2013multi}

\subsubsection{Hyper-parameters}CNNs' parameters are especially tricky to train, as it contains more hyper-parameters than a standard MLP (Multilayer perceptron). While the general rules of thumb for learning rates and regularization constants still apply, the following should be kept in mind when optimizing CNNs \cite{jaderberg2014synthetic}. Since the feature map size decreases with the depth of layers increasing, layers near the input layer will tend to have fewer filters, while layers higher up can have much more filters. In fact, to equalize computations at each layer, the product of the number of features and the number of pixel positions is typically picked to be roughly constant across layers. To preserve the information about the input, it is required to keep the total number of activations (number of feature maps times number of pixel positions) to be non-decreasing from one layer to the next. The number of feature maps directly controls the capacity, while the features are dependant on the number of available examples and the complexity of the task\cite{hinton2012improving}\cite{krizhevsky2012imagenet}.
\subsubsection{Datasets}Our datasets are mentioned in Fig. 2, which come from three major sources to provide CAPTCHA examples. The challenge is the insufficient size of datasets. However, with the increase of training data size, the performance of accuracy will correspondingly improve. We propose a method to gain a larger data size as shown in Fig. 10. Through careful observation, we found that the majority of rotation angels in examples varying from $-50^{\circ}$ to $+50^{\circ}$, but our manually training samples are insufficient which means cannot cover the comprehensively rotated CAPTCHA examples. Thus we rotate our data every  $5^{\circ}$ to get the complete samples.

\begin{figure}
\centering
\includegraphics[width =0.65\textwidth]{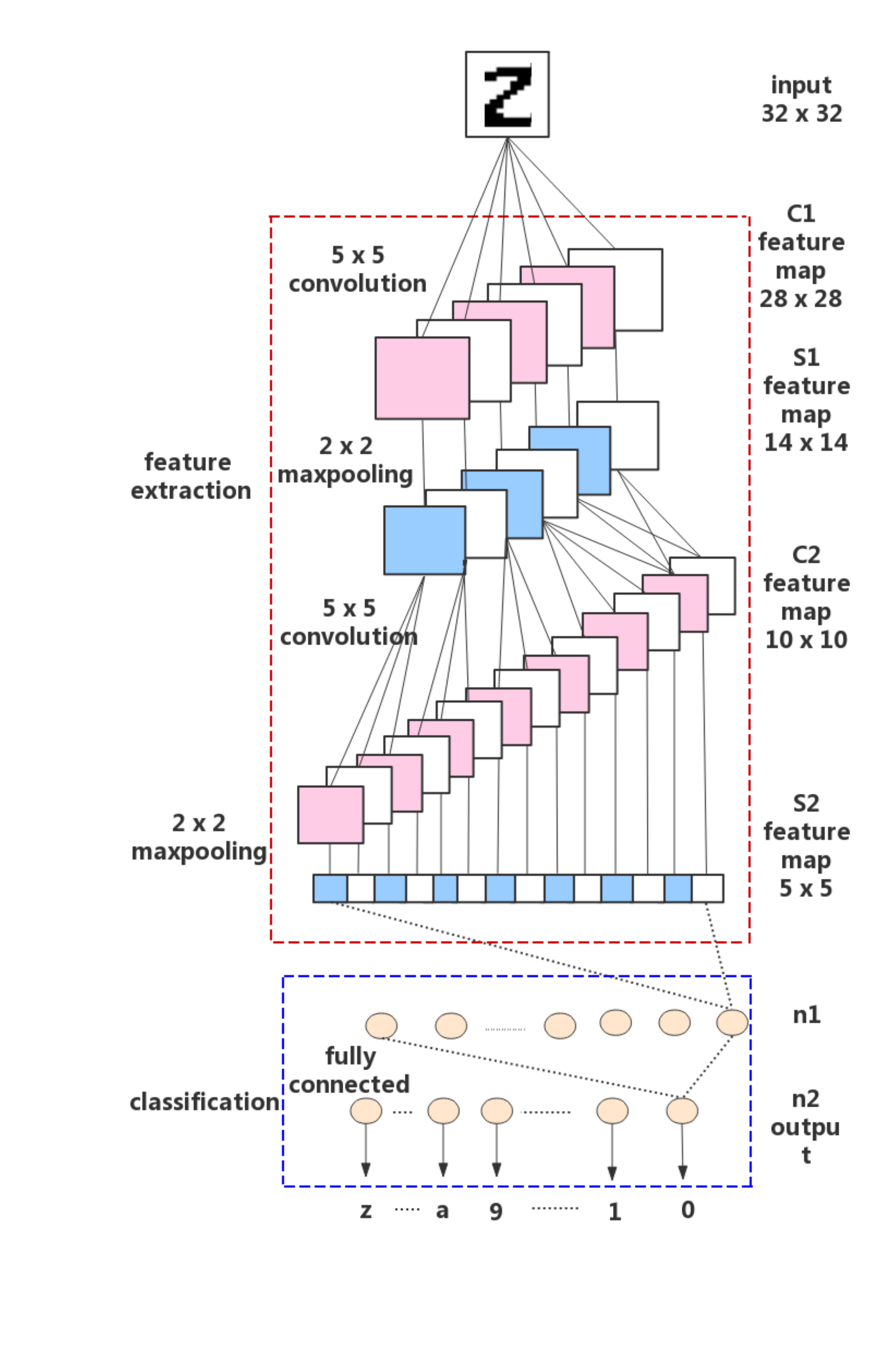}
\caption{Neural network structure}\label{fig9}
\end{figure}

\begin{figure}
\centering
\includegraphics[width = 0.35\textwidth]{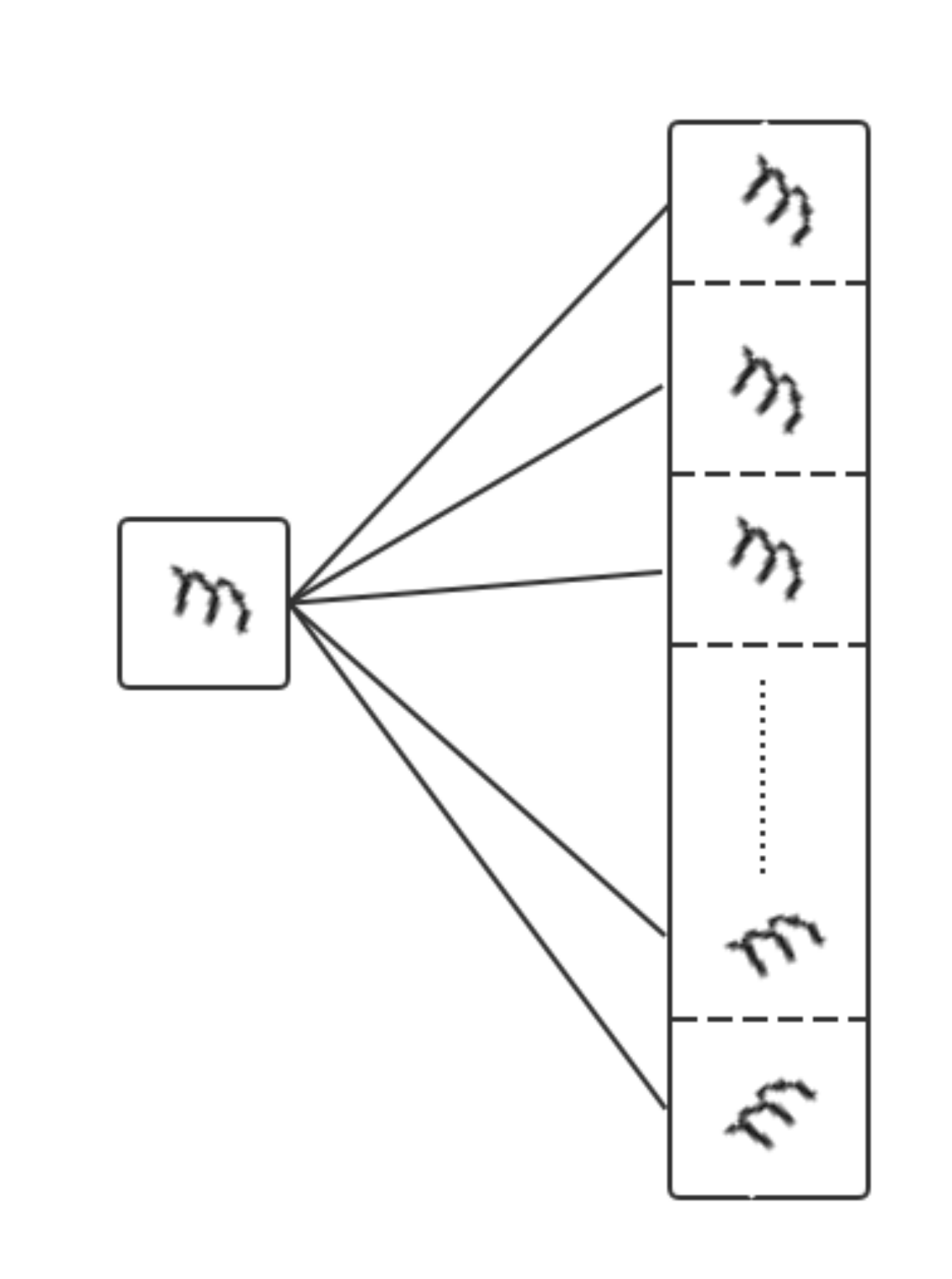}
\caption{Rotate the image to diversify training data}\label{fig10}
\end{figure}

\section{Experiment}

More than 7,000 testing samples were selected in our experiment work. The whole process of defeating the CAPTCHA is illustrated in Fig. 11 including denoising, segmentation and recognition. We present the accuracy rate results by our methods in Fig. 12. Overall, CNN is much more robust than OCR/TM. However, the way to obtain the training data is really painful because we have to manually decompose the examples into single characters and label them. Besides, we rotate the training data from $-50^{\circ}$ to $50^{\circ}$ in every $5^{\circ}$ for enlarging the data size ten times more than before. Rotating the training sample also brings more comprehensively circumstances for potential testing data, which proved to be more useful. This is because our raw data size is limited and cannot cover the whole rotated cases. However, we can fulfill the rotation completeness by our method.

Shown in Fig. 12 are the results for comparison. It can be seen that the recognition rate for the particularly rotated CAPTCHAs such as the 9th and 10th row are much lower than the others. The best rotated recognition rate is 71.6\% in the sixth row in Fig. 12, and the lowest accuracy rate of rotated fonts is as low as 33\%. With the increasing rotated angle, the accuracy rate is correspondingly decreasing. Therefore, the rotation hinders the accuracy of segmentation and recognition. 
However, CNN can always perform much more robust than TM/OCR especially in rotation types. The average performance in CNN is approximately 10\% better than in TM/OCR. 

In fact, in some simple cases like first row show in Fig. 12, both TM/OCR and CNN perform similar in accuracy rate. With the rotated angles increase, more advantages appear by CNN than TM/OCR. Fig. 13 shows the accuracy rate much clearer. 

Our CAPTCHA examples are always constructed by four characters. If and only if all of them have been correctly recognized, we consider it as a successful defeat as we introduced before. More precisely, Fig. 14 shows the single character recognition accuracy rate. As we can see, the lowest accuracy rate is 75.96\% in TM/OCR and 85.72\% in CNN. The highest accuracy rate in Fig. 14 is 98.98\% in TM/OCR and 99.65\% in CNN, which indicates that our method has achieved the state-of-the-art performance.

\begin{figure}
\centering
\includegraphics[width = 0.65\textwidth]{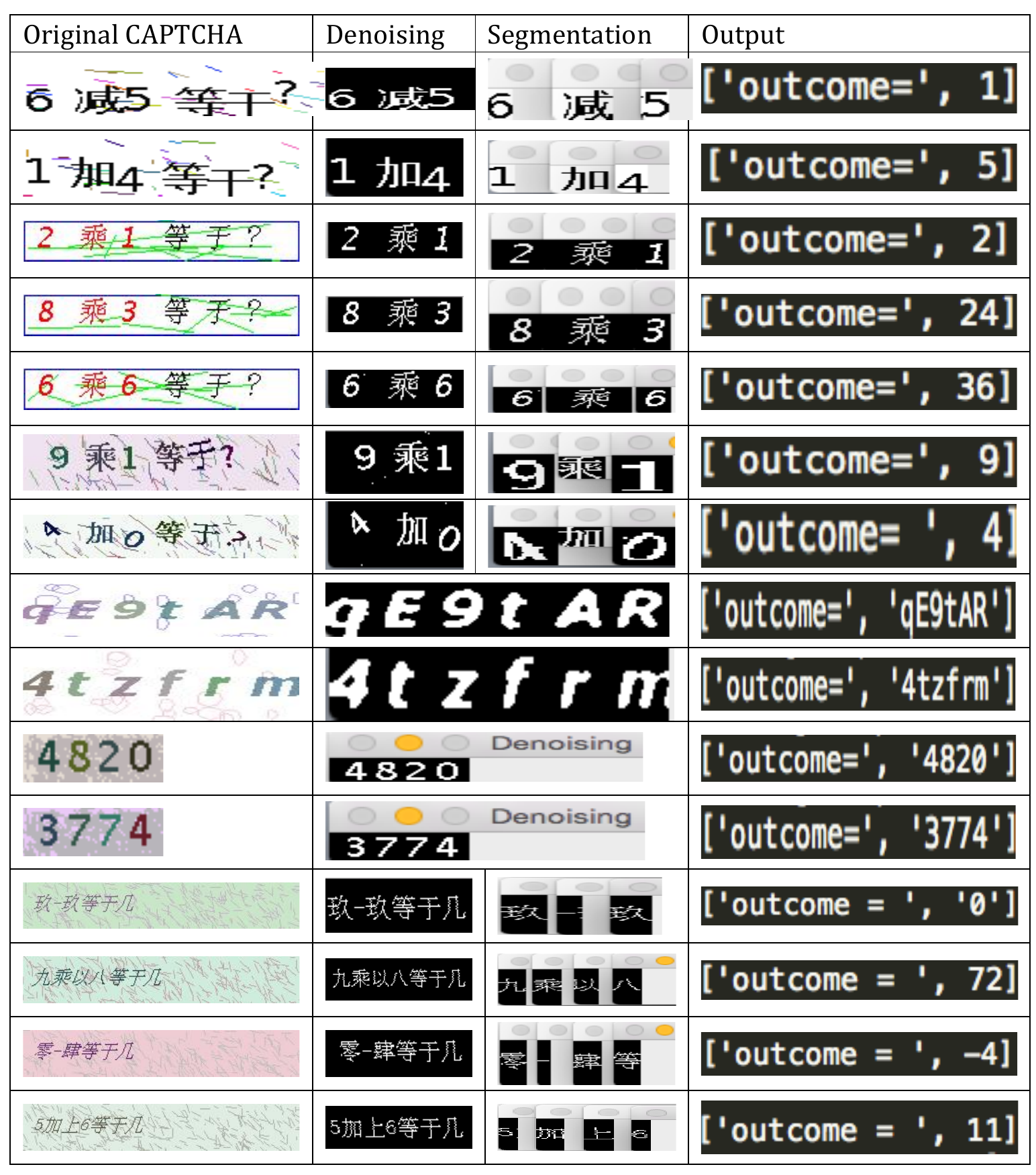}
\caption{The full process of defeating the CAPTCHA}\label{fig11}
\end{figure}

\begin{figure}
\centering
\includegraphics[width = 0.65\textwidth]{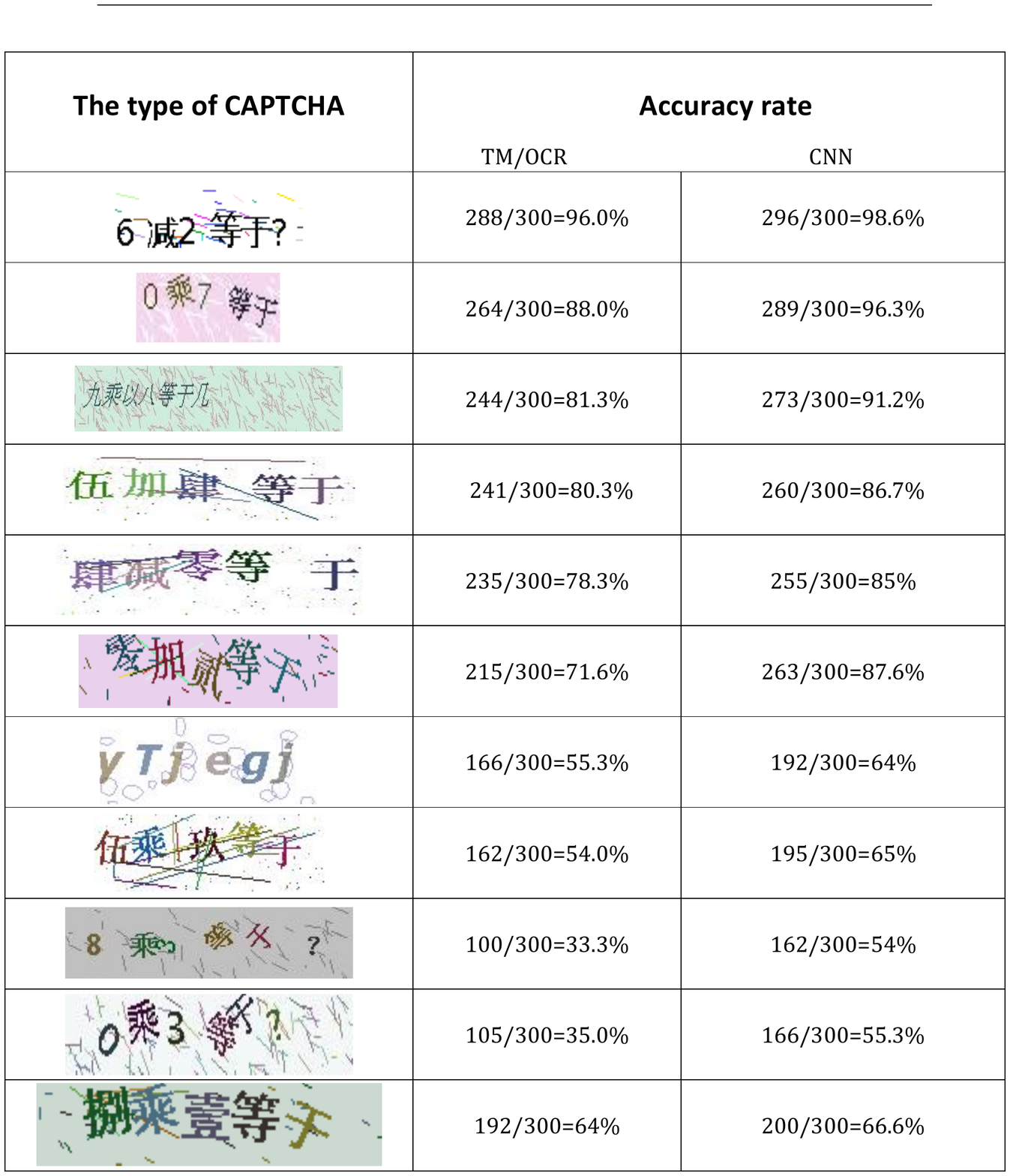}
\caption{The accuracy rate for each CAPTCHA}\label{fig12}
\end{figure}

\begin{figure}
\centering
\includegraphics[width = 0.65\textwidth]{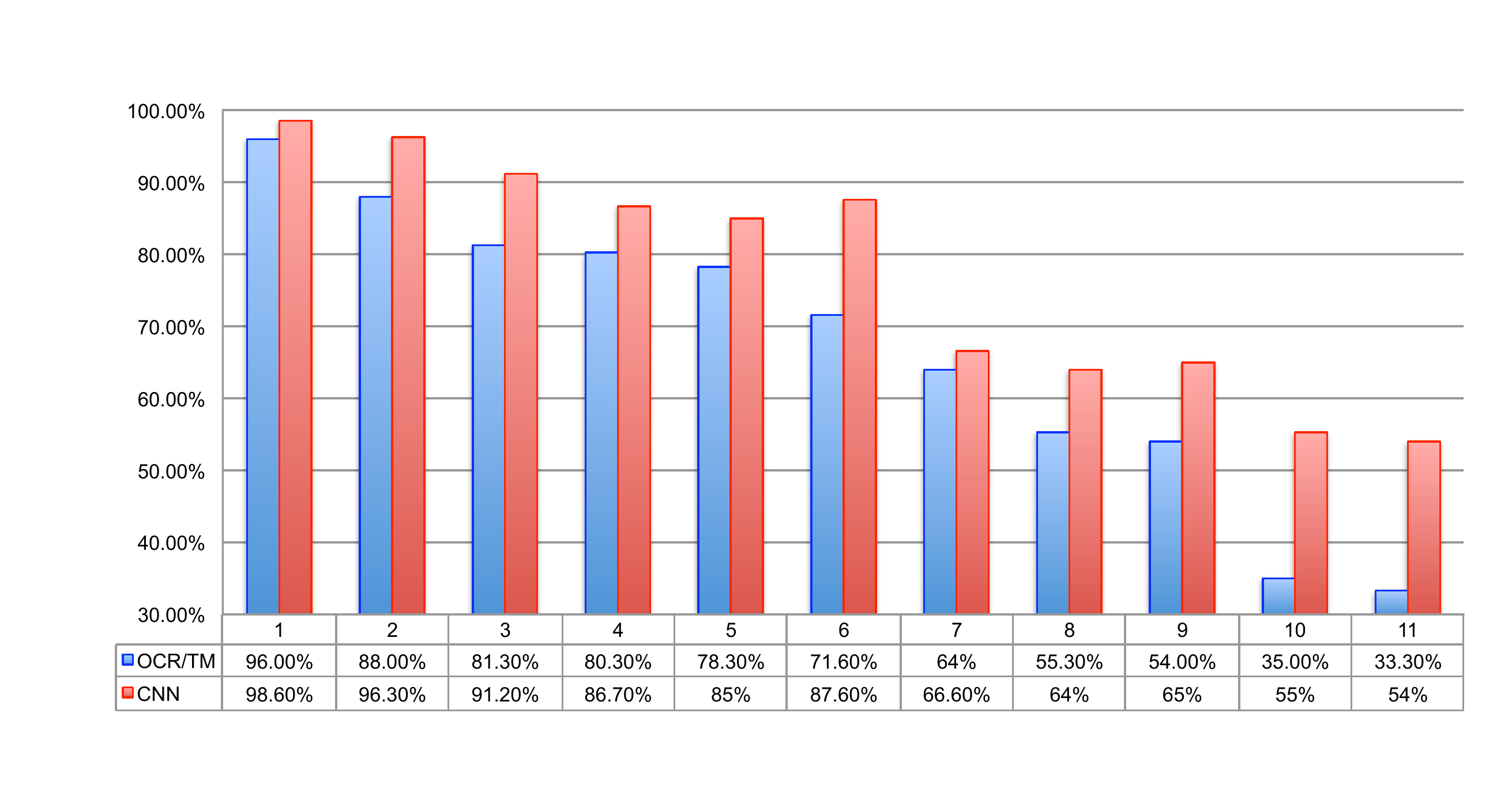}
\caption{Overall CAPTCHA accuracy rate}\label{fig13}
\end{figure}

\begin{figure}
\centering
\includegraphics[width = 0.65\textwidth]{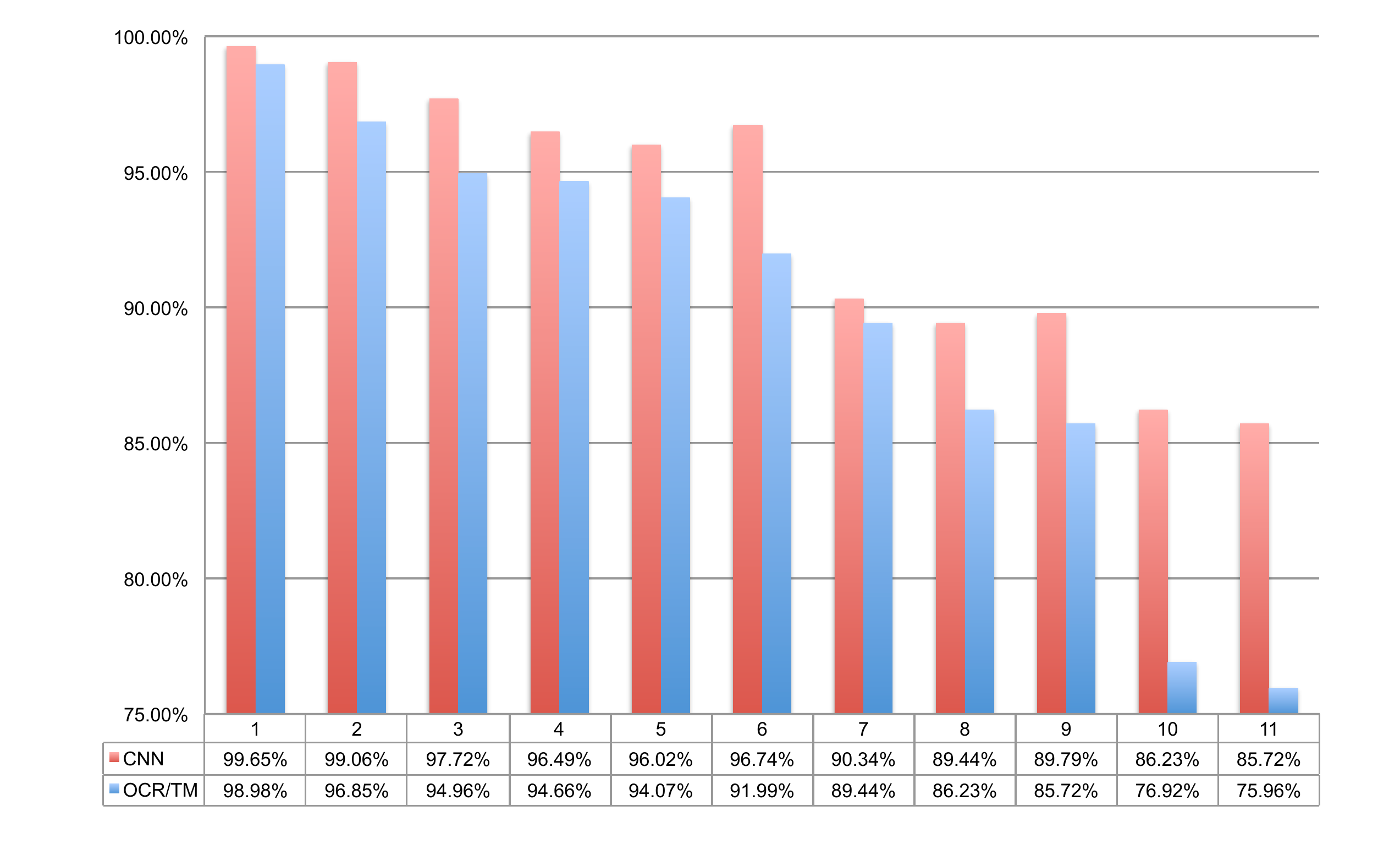}
\caption{Accuracy rate of individual character}\label{fig14}
\end{figure}

\section{Conclusions}
We created our optimized system including an adaptive length system and an optimal classifier to achieve the state-of-the-art performance in three major datasets (State Administration for Industry \& Commerce, Credit
Reference Center and State Intellectual Property Office). 
The adaptive system can perform 10.67 times faster than the traditional sequential segmentation, and the successful single character recognition rate is improved variously from 75\% to 99\% depending on the type of CAPTCHA.

Defeating the CAPTCHAs is also beneficial to improving the safety when we expose the CAPTCHAs' deficiency. Regarding the recognition performance, although TM/OCR is effective, fast and cost-effective, we can only use OCR in ideal condition, such as twisted-free, noise-free and rotation-free. This is because with the situation getting more sophisticated, the performance in TM/OCR drops dramatically. However, CNN acts as a more reliable classifier than TM/OCR during the whole test. 

In conclusion, CNN cannot perform well without sufficient high quality and quantity training data and reliable neural network. The cost to collect the training data in the first period is huge. We also proposed our method to reduce the collecting cost massively. Another contribution of our paper is completely solving the CAPTCHAs, in terms of manually collecting sufficient training data, evaluating different existing methods in practice, and combining the optimal method with our proposed algorithm to achieve the state-of-the-art performance.

 
\section{References}

\bibliographystyle{splncs03}
\bibliography{myreference}

\begin{thebibliography}{10}
\providecommand{\url}[1]{#1}
\csname url@samestyle\endcsname
\providecommand{\newblock}{\relax}
\providecommand{\bibinfo}[2]{#2}
\providecommand{\BIBentrySTDinterwordspacing}{\spaceskip=0pt\relax}
\providecommand{\BIBentryALTinterwordstretchfactor}{4}
\providecommand{\BIBentryALTinterwordspacing}{\spaceskip=\fontdimen2\font plus
\BIBentryALTinterwordstretchfactor\fontdimen3\font minus
  \fontdimen4\font\relax}
\providecommand{\BIBforeignlanguage}[2]{{%
\expandafter\ifx\csname l@#1\endcsname\relax
\typeout{** WARNING: IEEEtran.bst: No hyphenation pattern has been}%
\typeout{** loaded for the language `#1'. Using the pattern for}%
\typeout{** the default language instead.}%
\else
\language=\csname l@#1\endcsname
\fi
#2}}
\providecommand{\BIBdecl}{\relax}
\BIBdecl

\bibitem{bursztein2011text}
E.~Bursztein, M.~Martin, and J.~Mitchell, ``Text-based captcha strengths and
  weaknesses,'' in \emph{Proceedings of the 18th ACM conference on Computer and
  communications security}.\hskip 1em plus 0.5em minus 0.4em\relax ACM, 2011,
  pp. 125--138.

\bibitem{chellapilla2005building}
K.~Chellapilla, K.~Larson, P.~Y. Simard, and M.~Czerwinski, ``Building
  segmentation based human-friendly human interaction proofs (hips),'' in
  \emph{Human Interactive Proofs}.\hskip 1em plus 0.5em minus 0.4em\relax
  Springer, 2005, pp. 1--26.

\bibitem{ling2012case}
X.~Ling-Zi and Z.~Yi-Chun, ``A case study of text-based captcha attacks,'' in
  \emph{Cyber-Enabled Distributed Computing and Knowledge Discovery (CyberC),
  2012 International Conference on}.\hskip 1em plus 0.5em minus 0.4em\relax
  IEEE, 2012, pp. 121--124.

\bibitem{perreault2007median}
S.~Perreault and P.~H{\'e}bert, ``Median filtering in constant time,''
  \emph{Image Processing, IEEE Transactions on}, pp. 2389--2394, 2007.

\bibitem{wang2016self}
Y.~Wang and M.~Lu, ``A self-adaptive algorithm to defeat text-based captcha,''
  in \emph{2016 IEEE International Conference on Industrial Technology
  (ICIT)}.\hskip 1em plus 0.5em minus 0.4em\relax IEEE, 2016, pp. 720--725.

\bibitem{huang2010efficient}
S.-Y. Huang, Y.-K. Lee, G.~Bell, and Z.-h. Ou, ``An efficient segmentation
  algorithm for captchas with line cluttering and character warping,''
  \emph{Multimedia Tools and Applications}, vol.~48, no.~2, pp. 267--289, 2010.

\bibitem{smith2007overview}
R.~Smith, ``An overview of the tesseract ocr engine,'' in \emph{icdar}.\hskip
  1em plus 0.5em minus 0.4em\relax IEEE, 2007, pp. 629--633.

\bibitem{wang2017combining}
Y.~Wang, Y.~Huang, W.~Zheng, Z.~Zhou, D.~Liu, and M.~Lu, ``Combining
  convolutional neural network and self-adaptive algorithm to defeat synthetic
  multi-digit text-based captcha,'' in \emph{Industrial Technology (ICIT), 2017
  IEEE International Conference on}.\hskip 1em plus 0.5em minus 0.4em\relax
  IEEE, 2017, pp. 980--985.

\bibitem{lewis1995fast}
J.~P. Lewis, ``Fast template matching,'' in \emph{Vision interface}, vol.~95,
  no. 120123, 1995, pp. 15--19.

\bibitem{krizhevsky2012imagenet}
A.~Krizhevsky, I.~Sutskever, and G.~E. Hinton, ``Imagenet classification with
  deep convolutional neural networks,'' in \emph{Advances in neural information
  processing systems}, 2012, pp. 1097--1105.

\bibitem{lawrence1997face}
S.~Lawrence, C.~L. Giles, A.~C. Tsoi, and A.~D. Back, ``Face recognition: A
  convolutional neural-network approach,'' \emph{IEEE transactions on neural
  networks}, vol.~8, no.~1, pp. 98--113, 1997.

\bibitem{von2003captcha}
L.~Von~Ahn, M.~Blum, N.~J. Hopper, and J.~Langford, ``Captcha: Using hard ai
  problems for security,'' in \emph{Advances in Cryptology---EUROCRYPT
  2003}.\hskip 1em plus 0.5em minus 0.4em\relax Springer, 2003, pp. 294--311.

\bibitem{roshanbin2013survey}
N.~Roshanbin and J.~Miller, ``A survey and analysis of current captcha
  approaches,'' \emph{Journal of Web Engineering}, vol.~12, no. 1-2, pp. 1--40,
  2013.

\bibitem{yan2008low}
J.~Yan and A.~S. El~Ahmad, ``A low-cost attack on a microsoft captcha,'' in
  \emph{Proceedings of the 15th ACM conference on Computer and communications
  security}.\hskip 1em plus 0.5em minus 0.4em\relax ACM, 2008, pp. 543--554.

\bibitem{huang2008projection}
S.-Y. Huang, Y.-K. Lee, G.~Bell, and Z.-h. Ou, ``A projection-based
  segmentation algorithm for breaking msn and yahoo captchas,'' in \emph{Proc.
  of the international Conference of Signal and Image Engineering}.\hskip 1em
  plus 0.5em minus 0.4em\relax Citeseer, 2008.

\bibitem{liu2013efficient}
P.~Liu, J.~Shi, L.~Wang, and L.~Guo, ``An efficient ellipse-shaped blobs
  detection algorithm for breaking facebook captcha,'' in \emph{Trustworthy
  Computing and Services}.\hskip 1em plus 0.5em minus 0.4em\relax Springer,
  2013, pp. 420--428.

\bibitem{stearns1995method}
C.~C. Stearns and K.~Kannappan, ``Method for 2-d affine transformation of
  images,'' Dec.~12 1995, uS Patent 5,475,803.

\bibitem{rashidi2012implementation}
B.~Rashidi and B.~Rashidi, ``Implementation of a high speed technique for
  character segmentation of license plate based on thresholding algorithm,''
  \emph{International Journal of Image, Graphics and Signal Processing
  (IJIGSP)}, vol.~4, no.~12, pp. 9--18, 2012.

\bibitem{kurugollu2001color}
F.~Kurugollu, B.~Sankur, and A.~E. Harmanci, ``Color image segmentation using
  histogram multithresholding and fusion,'' \emph{Image and vision computing},
  pp. 915--928, 2001.

\bibitem{glasbey1993analysis}
C.~A. Glasbey, ``An analysis of histogram-based thresholding algorithms,''
  \emph{CVGIP: Graphical models and image processing}, pp. 532--537, 1993.

\bibitem{smith2009adapting}
R.~Smith, D.~Antonova, and D.-S. Lee, ``Adapting the tesseract open source ocr
  engine for multilingual ocr,'' in \emph{Proceedings of the International
  Workshop on Multilingual OCR}.\hskip 1em plus 0.5em minus 0.4em\relax ACM,
  2009, p.~1.

\bibitem{goodfellow2013multi}
I.~J. Goodfellow, Y.~Bulatov, J.~Ibarz, S.~Arnoud, and V.~Shet, ``Multi-digit
  number recognition from street view imagery using deep convolutional neural
  networks,'' \emph{arXiv preprint arXiv:1312.6082}, 2013.

\bibitem{lecun1998gradient}
Y.~LeCun, L.~Bottou, Y.~Bengio, and P.~Haffner, ``Gradient-based learning
  applied to document recognition,'' \emph{Proceedings of the IEEE}, vol.~86,
  no.~11, pp. 2278--2324, 1998.

\bibitem{rumelhart1988learning}
D.~E. Rumelhart, G.~E. Hinton, and R.~J. Williams, ``Learning representations
  by back-propagating errors,'' \emph{Cognitive modeling}, vol.~5, no.~3, p.~1,
  1988.

\bibitem{simard2003best}
P.~Y. Simard, D.~Steinkraus, J.~C. Platt \emph{et~al.}, ``Best practices for
  convolutional neural networks applied to visual document analysis.'' in
  \emph{ICDAR}, vol.~3, 2003, pp. 958--962.

\bibitem{jaderberg2014synthetic}
M.~Jaderberg, K.~Simonyan, A.~Vedaldi, and A.~Zisserman, ``Synthetic data and
  artificial neural networks for natural scene text recognition,'' \emph{arXiv
  preprint arXiv:1406.2227}, 2014.

\bibitem{hinton2012improving}
G.~E. Hinton, N.~Srivastava, A.~Krizhevsky, I.~Sutskever, and R.~R.
  Salakhutdinov, ``Improving neural networks by preventing co-adaptation of
  feature detectors,'' \emph{arXiv preprint arXiv:1207.0580}, 2012.

\end{thebibliography}

\end{document}